\documentclass[preprint,12pt]{elsarticle}

\usepackage{amssymb}
\usepackage{amsmath}
\usepackage{url}
\usepackage{graphicx}
\usepackage{array}
\usepackage{multirow}
\usepackage{booktabs}
\usepackage{tabularx}
\usepackage{makecell}
\usepackage{subcaption}
\usepackage[table]{xcolor}
\definecolor{prefblue}{HTML}{214761}
\definecolor{prefshade}{HTML}{EAF1F5}
\usepackage{enumitem}
\setlist[itemize]{itemsep=0pt,parsep=0pt,topsep=2pt,partopsep=0pt,leftmargin=*}
\setlist[enumerate]{itemsep=1pt,parsep=0pt,topsep=2pt,partopsep=0pt,leftmargin=*}
\usepackage{float}
\usepackage{placeins}
\usepackage{longtable}
\usepackage{pdflscape}
\usepackage{hyperref}
\usepackage{cleveref}
\usepackage{algorithm}
\usepackage{algpseudocode}
\hypersetup{
  hidelinks,
  pdftitle={PREF-Gate: Provenance-Constrained Relational Evidence Fusion with Validation-Gated Selection for Graph Fraud Detection},
  pdfauthor={Liming Liu; Chao Hu; Mingfei Lu; Yiwei Ge; Xingle Li; Heyuan Shi}
}

\journal{Knowledge-Based Systems}
\graphicspath{{./}}
\date{}
\makeatletter
\def\ps@pprintTitle{%
  \let\@oddhead\@empty
  \let\@evenhead\@empty
  \let\@oddfoot\@empty
  \let\@evenfoot\@empty}
\makeatother

\begin{document}
\begin{frontmatter}
\title{PREF-Gate: Provenance-Constrained Relational Evidence Fusion with Validation-Gated Selection for Graph Fraud Detection}

\author[1]{Liming Liu\corref{cor1}}
\ead{244512048@cs.edu.cn}
\author[1]{Chao Hu}
\author[2]{Mingfei Lu}
\author[1]{Yiwei Ge}
\ead{8206240605@csu.edu.cn}
\author[1]{Xingle Li}
\ead{8206240606@csu.edu.cn}
\author[1]{Heyuan Shi}
\cortext[cor1]{Corresponding author}

\affiliation[1]{organization={Central South University},
                city={Changsha},
                country={China}}
\affiliation[2]{organization={University of Technology Sydney,
                Australian Artificial Intelligence Institute},
                city={Sydney},
                country={Australia}}

\begin{abstract}
Relational fraud detection can exploit both label-free graph context and
label-derived neighborhood evidence, but these two information sources obey
different validity conditions. In particular, neighborhood risk becomes
invalid when a queried node's own label, or any validation or test label,
enters its construction. We formulate this issue as provenance-constrained
relational evidence use and present PREF-Gate, an auditable decision framework
with two fixed experts and a finite validation gate. The context expert uses
attributes, one-hop means, feature residuals, and degree descriptors without
labels. The evidence expert adds self-excluded, training-label-only
neighborhood risk and empirical-Bayes summaries that expose support,
uncertainty, availability, and shrinkage. Before test inference, the gate
selects either expert or one of three pre-specified probability mixtures and
fixes the decision threshold. On Amazon, YelpChi, and TFinance, using five
identical stratified splits and 14 same-protocol methods, PREF-Gate obtains mean
AUPRC values of 0.9085, 0.8104, and 0.8913. It selects the label-free expert on
all Amazon and YelpChi splits and an evidence mixture on all TFinance splits.
Thus, the main result is conditional rather than universal: label-derived
relational evidence is useful only where held-out validation supports it. The
framework couples competitive ranking performance with an explicit
label-provenance contract, finite selection policy, failure accounting, and
review-budget evaluation, providing an auditable knowledge-based decision
pipeline for graph fraud detection.
\end{abstract}

\begin{keyword}
graph fraud detection \sep relational evidence \sep label provenance \sep
validation-gated fusion \sep empirical Bayes \sep decision support
\end{keyword}

\end{frontmatter}

\section{Introduction}
\label{sec:introduction}

Fraud detection is an imbalanced decision problem embedded in a relational
system. A platform may inspect only a small fraction of users, reviews, or
transactions, while fraudulent entities share devices, products, counterparties,
and structural patterns. These relations provide contextual knowledge that is
not present in an entity's attributes. Graph neural networks (GNNs) offer one
way to learn from such dependencies
\cite{kipf2017gcn,hamilton2017graphsage,velickovic2018gat,xu2019gin}, but model
expressiveness alone does not determine whether relational information is
valid or useful for an operational decision.

The first unresolved issue is evidence provenance. One-hop feature context is
label free, whereas neighborhood fraud risk is knowledge derived from observed
labels. The latter can be highly predictive while being scientifically invalid
if validation or test labels enter its numerator, denominator, prior, or
selection rule. Even a training node can leak its target through a self-loop or
a global prior. Because transductive graph pipelines commonly load all nodes
and edges together, the distinction must be enforced at the evidence operator,
not left as an informal property of the data split.

The second issue is evidence reliability. A neighborhood rate
supported by many eligible labels is different from the same rate supported by
one label. On another dataset, raw attributes and label-free neighborhood
summaries may already be sufficiently discriminative, so adding label-derived
evidence can reduce generalization. A fixed fusion policy can therefore turn an
occasionally useful signal into a systematic source of variance.

This paper presents PREF-Gate, a provenance-constrained relational-evidence and
validation-gated decision framework for graph fraud detection. The method
contains two fixed experts. The
\emph{label-free context expert} uses raw node attributes, one-hop neighbor
means, absolute feature residuals, log degree, and an isolation indicator. The
\emph{evidence context expert} augments that representation with neighborhood
risk computed only from eligible training labels, together with
empirical-Bayes posterior mean and variance, support, availability, shrinkage
displacement, and the applicable training prevalence. A finite gate considers
the two experts and three pre-specified convex probability mixtures. It selects
one candidate using validation Macro-F1 and AUPRC, freezes a validation-selected
threshold, and only then requests test probabilities. Relational statistics
are explicit evidence objects with recorded provenance and reliability; the
gate decides whether they modify the label-free output.

\paragraph{Model identity}
PREF-Gate denotes the complete representation, evidence-construction, and
adaptive-gating procedure. ``AG'' refers to the pre-specified adaptive gate.
The selected predictor is not claimed to be an end-to-end neural
message-passing encoder: its experts are tree ensembles trained on explicit
graph-derived representations, while learned GNNs enter the comparison roster.

\begin{figure}[H]
  \centering
  \includegraphics[width=0.98\linewidth]{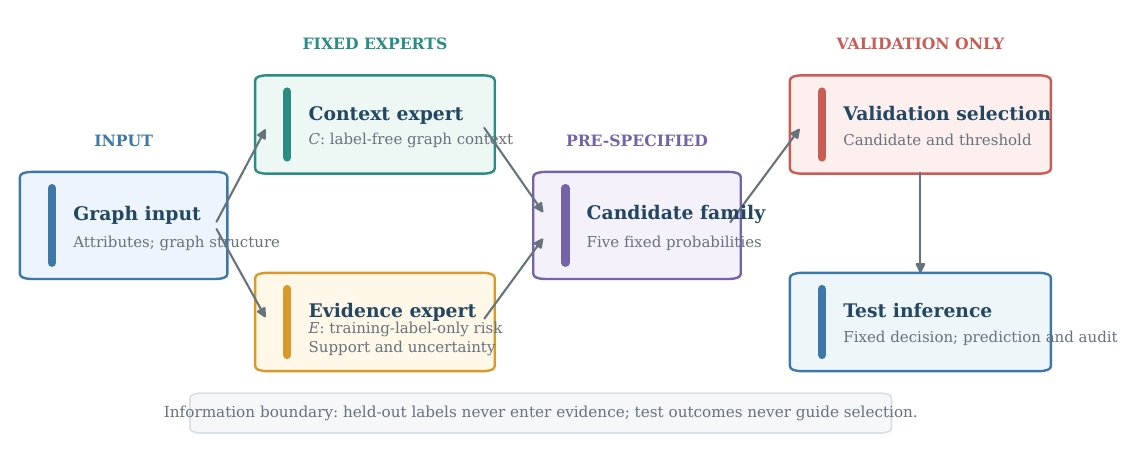}
  \caption{PREF-Gate evidence and decision workflow. Label-free context and
  training-label-only evidence have separate provenance paths. The five
  candidate decisions are specified in advance; validation chooses a candidate
  and threshold before test probabilities are requested.}
  \label{fig:overview}
\end{figure}

The adaptive gate is a deliberately conservative revision of an exploratory
35-candidate gate. The larger gate compared five experts and 30 pairwise
mixtures; its validation advantage was sometimes as small as $6\times10^{-5}$,
which was insufficient to justify switching away from a stable context expert.
The final gate reduces the candidate family to the scientific question of
interest: whether training-label evidence should modify a strong label-free
graph context predictor. This change is not an unconstrained ensemble. The
experts, mixture grid, selection metric, split files, and threshold rule are
all finite and recorded before the adaptive-gate execution.

The evaluation uses Amazon, YelpChi, and TFinance with five identical
stratified 40\%/20\%/40\% train/validation/test splits for every method. The
roster contains feature and graph baselines, PREF-Gate component ablations,
CARE-GNN \cite{dou2020caregnn}, PC-GNN \cite{liu2021pcgnn}, BWGNN
\cite{tang2022bwgnn}, and two recent ICLR 2024 methods: PMP
\cite{zhuo2024partitioning} and ConsisGAD \cite{chen2024consistency}. Across the
three datasets, the adaptive gate obtains mean AUPRC values of 0.9085, 0.8104,
and 0.8913. These values exceed the strongest successfully completed external
baseline by 0.0389, 0.0764, and 0.0245, respectively.

On Amazon and YelpChi, PREF-Gate converges to the label-free graph-context expert,
so the gate rejects unsupported label-derived evidence. On TFinance it selects
a context/evidence mixture in every split. The gate therefore admits evidence
only when validation supports it; it does not guarantee additive gains on
every dataset.

The contributions are:
\begin{itemize}
  \item We formalize a label-provenance contract for relational evidence,
  including self-exclusion, leave-one-out training priors, and a strict
  separation between evidence construction, validation decisions, and test
  evaluation.
  \item We construct an auditable empirical-Bayes evidence object that keeps
  risk magnitude, support, posterior uncertainty, availability, and shrinkage
  displacement distinct rather than reducing them to one neighborhood rate.
  \item We introduce a finite validation-gated decision rule that conditionally
  admits label-derived evidence while retaining a label-free context expert as
  a defined outcome, thereby exposing negative evidence rather than concealing
  it behind a fixed ensemble.
  \item We evaluate the complete decision pipeline under identical splits,
  review budgets, calibration measures, paired tests, and explicit execution
  accounting against 13 baselines and ablations, including two ICLR 2024
  methods.
\end{itemize}

The remainder of the paper reviews the closest graph-fraud and evidence
literature, formalizes the information boundary, presents PREF-Gate, and then
describes the experimental protocol and research questions. The final sections
report the corresponding results, limitations, and reproducibility record.

\section{Related Work}
\label{sec:related}

\subsection{Graph fraud detection}

Fraud-specific GNNs address properties that generic message passing does not
explicitly model. CARE-GNN learns relation-aware neighbor selection to reduce
camouflage \cite{dou2020caregnn}. PC-GNN combines label-balanced sampling and
neighbor selection for severe class imbalance \cite{liu2021pcgnn,he2009learning}.
GraphSMOTE synthesizes minority representations and links
\cite{zhao2021graphsmote}, while SemiGNN combines supervised and
semi-supervised objectives \cite{Wang2019}. GraphConsis models feature, label,
and relation inconsistency \cite{Liu2020SIGIR}; FraudRE jointly considers
inconsistency and imbalance \cite{Zhang2021}; and H2-FDetector separates
homophilic and heterophilic connections \cite{Shi2022}.

Spectral and heterophily-aware approaches provide another view. BWGNN uses
band-pass filters to expose anomalous graph frequencies
\cite{tang2022bwgnn}. GHRN analyzes graph anomaly detection through a spectral
heterophily perspective \cite{Gao2023}, and AO-GNN adapts optimization to the
fraud task \cite{Huang2022}. DiG-In-GNN uses discriminative feature guidance in
inconsistent multi-relation fraud graphs \cite{zhang2024digingnn}. These
methods motivate strong graph-aware comparisons, but they differ in label
access, splitting conventions, and supported datasets. A number copied from a
paper table is not equivalent to a result generated under the present split
and leakage boundary.

\subsection{Recent graph-fraud and anomaly baselines}

PMP partitions neighbor messages by class-related roles so a majority benign
population does not dominate aggregation \cite{zhuo2024partitioning}.
ConsisGAD combines consistency training with learnable augmentation under
limited supervision \cite{chen2024consistency}. Both provide public source and
could be adapted to caller-supplied split indices while deferring test
inference, so they enter the quantitative comparison.

Other 2024--2025 methods were screened but not forced into the Cartesian
benchmark. A method was eligible only if its public implementation could cover
Amazon, YelpChi, and TFinance, consume the exact split indices, prevent
non-training labels from entering training, select checkpoints on validation,
and defer test inference. Methods with a different task definition,
incomplete dataset coverage, unavailable dependency stack, or a stock loop
that repeatedly evaluates test data are discussed as related work rather than
assigned an invented number. This distinction restricts the quantitative
comparison to evidence that can be rerun fairly.

The 2025 literature further broadens the design space. GAAP globally aggregates
attribute-association patterns \cite{duan2025gaap}; CGNN targets graph fraud
detection with extremely limited labels \cite{li2025cgnn}; MimbFD addresses
topological and class-driven message imbalance through dual-view representation
learning \cite{song2025mimbfd}; and HUGE studies label-free unsupervised graph
fraud detection \cite{pan2025huge}. These are relevant contemporary methods,
but their released supervision settings, available dataset configurations, or
local adapter status do not form the same prespecified 40/20/40 supervised Cartesian
matrix. They are therefore cited and screened rather than assigned
non-comparable numbers.

\subsection{Graph anomaly detection and context features}

Attributed-graph anomaly methods include reconstruction, contrastive, and
spectral objectives. DOMINANT reconstructs attributes and topology
\cite{ding2019dominant}; CoLA learns agreement between nodes and local
subgraphs \cite{liu2022cola}; and broader analyses question when abnormality is
detectable by a GNN at all \cite{Chai2022}. These studies show that topology is
not automatically beneficial. When raw attributes are strong, neighborhood
summaries can act as informative context without requiring a deep learned
message-passing stack.

The label-free expert in this work uses a deterministic one-hop mean operator.
It is related to the aggregation step of GraphSAGE
\cite{hamilton2017graphsage}, but it materializes the aggregate as a design
matrix for a tree-based classifier. This makes the exact information flow easy
to audit and separates the value of graph context from the optimization
behavior of a neural encoder.

This distinction also limits the architecture claim. The empirical
contribution is not a new universal message-passing layer. It is the
combination of an auditable graph-context representation, self-excluded
relational evidence, and validation-only routing under a uniform benchmark.

\subsection{Label-derived features and empirical Bayes}

Neighborhood fraud rates resemble target encoding: the feature is predictive
precisely because it summarizes observed labels. Target statistics therefore
require an explicit source boundary \cite{micci2001target}. Ordered target
statistics in CatBoost illustrate the more general principle that a target
must not encode itself \cite{prokhorenkova2018catboost}. In a graph, exclusion
must account for self-loops and for a node's contribution to any global
training prior.

Empirical-Bayes shrinkage separates an observed rate from its uncertainty. A
Beta-binomial representation provides a posterior mean and variance whose
behavior depends on the number of eligible observations
\cite{wilcox1981betabinomial}. PREF-Gate does not replace raw risk with the
posterior mean. It retains raw risk, posterior summaries, support, availability,
and displacement so the evidence expert can learn how much trust to place in
the local rate.

\subsection{Calibration and constrained review}

AUPRC is appropriate for highly imbalanced ranking because it emphasizes
precision and recall for the positive class \cite{davis2006relationship}.
Operational fraud teams also work at finite review budgets, making
Precision@5\% and Recall@5\% directly interpretable. Probability calibration
is a separate property: a model can rank well while producing biased
probabilities \cite{guo2017calibration}. Graph dependence can further alter
calibration behavior \cite{hsu2022gats}. We therefore report Brier score and
10-bin expected calibration error (ECE), but do not conflate them with AUPRC.

\section{Problem Setting and Information Boundary}
\label{sec:setting}

Let $G=(V,E,X)$ be a graph with $n=|V|$ nodes, normalized homogeneous edge
set $E$, feature matrix $X\in\mathbb{R}^{n\times d}$, and binary labels
$y_i\in\{0,1\}$. For outer split $s$, the node memberships are
$\mathcal{T}_s$, $\mathcal{V}_s$, and $\mathcal{Q}_s$ for training,
validation, and test. The three sets are disjoint within a split.

All methods receive the same node features, normalized adjacency, and split
indices. Training labels $y_j$ for $j\in\mathcal{T}_s$ may train a classifier
or construct evidence. Validation labels may select a candidate, checkpoint,
mixture, and decision threshold. Test labels are available only to the final
metric evaluator.

\subsection{Relational evidence as provenance-typed knowledge}

We distinguish three information types. Attribute knowledge
$K_X=(X,E)$ is available without labels and may be used by every split.
Training-derived relational knowledge $K_R^{(s)}$ consists of statistics whose
label sources belong exclusively to $\mathcal{T}_s$. Decision knowledge
$K_D^{(s)}$ consists of validation scores, the selected candidate, and its
threshold; it may control routing but may not alter $K_R^{(s)}$. Test labels
belong only to the evaluator and to none of these knowledge objects.

This typing makes the validity condition compositional. A representation is
admissible for split $s$ only if each label-dependent component can be traced
to $K_R^{(s)}$ and each adaptive choice can be traced to $K_D^{(s)}$. The
following contract instantiates that condition for neighborhood evidence.

\subsection{Label-access contract}

For node $i$, the eligible labeled neighborhood is
\begin{equation}
\widetilde{\mathcal{N}}_s(i)=
\{j\in\mathcal{N}(i)\cap\mathcal{T}_s: j\neq i\}.
\label{eq:eligible}
\end{equation}
The $j\neq i$ condition is mandatory even when $i$ is a training node and the
normalized graph contains a self-loop. Validation and test nodes can contribute
their raw attributes and structural position in the transductive graph, but
their labels cannot contribute to evidence.

If the training prevalence is estimated automatically, a training node
receives a leave-one-out prevalence:
\begin{equation}
\pi_{s,-i}=
\frac{\sum_{j\in\mathcal{T}_s}y_j-y_i}
{|\mathcal{T}_s|-1},
\qquad i\in\mathcal{T}_s,
\label{eq:loo-prior}
\end{equation}
whereas validation and test nodes use the complete training prevalence
$\pi_s$. This prevents a node's target from entering its own evidence through
the smoothing prior.

\subsection{Prediction and review-budget objectives}

Each method produces a fraud probability $p_i$. A threshold $\tau_s$ is
selected on $\mathcal{V}_s$ by maximizing validation Macro-F1 and is frozen
before test inference. At review fraction $b=0.05$, the review set consists of
the top $\lceil b|\mathcal{Q}_s|\rceil$ test probabilities. We report its
precision and recall, together with AUPRC, Macro-F1, AUROC, Brier score, and
ECE.

\section{PREF-Gate: Evidence Construction and Decision Fusion}
\label{sec:method}

\subsection{Graph normalization}

Each released graph is converted to a homogeneous representation. For every
edge $(u,v)$, the reverse edge $(v,u)$ is added; duplicates are removed in a
deterministic lexical order; and one self-loop is inserted per node. Missing
feature values are mapped to zero, with infinite values clipped to finite
sentinels. The same normalized graph is passed to every independent method and
to external adapters that require homogeneous adjacency.

Let
\begin{equation}
\overline{x}_i =
\frac{1}{|\mathcal{N}(i)|}
\sum_{j\in\mathcal{N}(i)}x_j
\label{eq:mean}
\end{equation}
denote the incoming one-hop feature mean, with a zero vector for an empty
neighborhood. Because the normalized graph includes self-loops, the operator
has a stable value for every node.

\subsection{Label-free context expert}

The label-free context vector is
\begin{equation}
z_i^{(C)} =
\left[
x_i \,\|\, \overline{x}_i \,\|\,
|x_i-\overline{x}_i| \,\|\,
\log(1+\deg(i)) \,\|\,
\mathbb{I}[\deg(i)>0]
\right].
\label{eq:context}
\end{equation}
It captures three complementary views: the node itself, the average local
context, and the magnitude of disagreement between the two. The final two
terms expose neighborhood scale and availability.

A histogram gradient-boosting classifier $f_C$ is fit on
$\{(z_i^{(C)},y_i):i\in\mathcal{T}_s\}$. The fixed configuration uses 180
iterations, learning rate 0.05, 31 maximum leaf nodes, $L_2$ regularization
1.0, balanced class weighting, and random state $s+303$. Its validation and
test probabilities are denoted $p_i^{(C)}$.

\subsection{Training-label-only neighbor risk}

For eligible support
\begin{equation}
m_i = |\widetilde{\mathcal{N}}_s(i)|,\qquad
k_i = \sum_{j\in\widetilde{\mathcal{N}}_s(i)} y_j,
\end{equation}
the raw neighborhood risk is
\begin{equation}
r_i =
\begin{cases}
k_i/m_i, & m_i>0,\\
\pi_{s,-i}, & m_i=0,\ i\in\mathcal{T}_s,\\
\pi_s, & m_i=0,\ i\notin\mathcal{T}_s.
\end{cases}
\label{eq:risk}
\end{equation}
A separately reported smoothed risk is
\begin{equation}
\widetilde{r}_i =
\frac{k_i+\alpha\pi_i}{m_i+\beta},
\qquad \alpha=1,\quad \beta=2,
\label{eq:smoothed}
\end{equation}
where $\pi_i$ is the applicable self-excluded or complete training prevalence.
Neither validation nor test labels enter $k_i$, $m_i$, or $\pi_i$.

\subsection{Empirical-Bayes evidence}

The concentration $\kappa_s$ is estimated from supported training nodes. Let
$q_i=k_i/m_i$ for $i\in\mathcal{T}_s$ with $m_i>0$. A
support-weighted mean and variance are computed, and the expected binomial
sampling noise is subtracted. The resulting method-of-moments concentration is
clipped to $[2,100]$; when fewer than two supported training nodes exist, the
fallback concentration is 12.

For node-specific prior shapes
\begin{equation}
a_i=\kappa_s\pi_i,\qquad b_i=\kappa_s(1-\pi_i),
\end{equation}
the posterior summaries are
\begin{align}
\mu_i &=
\frac{k_i+a_i}{m_i+a_i+b_i},\\
\sigma_i^2 &=
\frac{(k_i+a_i)(m_i-k_i+b_i)}
{(m_i+a_i+b_i)^2(m_i+a_i+b_i+1)}.
\label{eq:posterior}
\end{align}
The evidence vector is
\begin{equation}
e_i =
\left[
r_i,\mu_i,\sigma_i^2,\log(1+m_i),
\mathbb{I}[m_i>0],r_i-\mu_i,\pi_i
\right].
\label{eq:evidence}
\end{equation}
The displacement $r_i-\mu_i$ records how strongly the posterior pulls the raw
rate toward the training prior.

\subsection{Evidence context expert}

The evidence expert receives
\begin{equation}
z_i^{(E)}=[z_i^{(C)}\|e_i].
\end{equation}
An Extra-Trees classifier $f_E$ with 200 trees, minimum leaf size 10, balanced
class weights, and random state $s+404$ is trained on $\mathcal{T}_s$. Its
probability is $p_i^{(E)}$. The model is deliberately different from $f_C$ so
that the evidence path can learn non-linear interactions among raw risk,
support, and posterior uncertainty.

\subsection{Finite validation-adaptive gate}

The pre-specified candidate family is
\begin{equation}
\mathcal{P}=
\left\{
p^{(C)},p^{(E)},
wp^{(C)}+(1-w)p^{(E)}:
w\in\{0.25,0.50,0.75\}
\right\}.
\label{eq:candidates}
\end{equation}
There are exactly five candidates. For each candidate $c$, validation
Macro-F1 determines a candidate-specific threshold $\tau_c$. The gate score is
\begin{equation}
S_c =
\frac{1}{2}\operatorname{MacroF1}_{\mathcal{V}_s}(c,\tau_c)
+
\frac{1}{2}\operatorname{AUPRC}_{\mathcal{V}_s}(c).
\label{eq:gate-score}
\end{equation}
Candidates are ordered by $S_c$, then validation AUPRC, validation Macro-F1,
and finally candidate name. The selected candidate $c_s^\star$ and threshold
$\tau_s^\star$ are frozen before either expert is asked for test
probabilities.

\begin{algorithm}[t]
\caption{Validation-adaptive PREF-Gate inference}
\label{alg:bpm}
\begin{algorithmic}[1]
\Require Graph $G$, features $X$, split
$(\mathcal{T}_s,\mathcal{V}_s,\mathcal{Q}_s)$
\State Build $z^{(C)}$ from $X$ and adjacency only
\State Build $e$ from labels in $\mathcal{T}_s$ with self-exclusion
\State Fit $f_C$ on $z^{(C)}_{\mathcal{T}_s}$ and $f_E$ on
$[z^{(C)}\|e]_{\mathcal{T}_s}$
\State Materialize validation probabilities for the two experts
\State Enumerate the five candidates in Eq.~\eqref{eq:candidates}
\For{each validation candidate $c$}
  \State Select $\tau_c$ by validation Macro-F1
  \State Compute $S_c$ using Eq.~\eqref{eq:gate-score}
\EndFor
\State Freeze $c_s^\star=\arg\max_c S_c$ and $\tau_s^\star$
\State Materialize only the required test expert probabilities
\State \Return selected test probabilities and frozen threshold
\end{algorithmic}
\end{algorithm}

\subsection{Why the gate is conservative}

The earlier full gate included structural, raw risk, empirical-Bayes risk,
context, and evidence experts plus every pairwise mixture at three weights.
With five base experts this produced 35 candidates. Maximizing over many
correlated validation scores creates a multiple-comparison effect: an
insignificant validation fluctuation can change the chosen model. The reduced
gate asks a narrower mechanism question and avoids unrelated weak candidates.

This design also permits an interpretable fallback. If evidence is not
supported, $p^{(C)}$ remains a valid final model rather than an emergency
exception. If evidence improves validation, a pre-specified mixture can be
selected. The fallback is therefore part of PREF-Gate's method definition.

\subsection{Complexity}

Constructing one-hop means and evidence is $O(|E|d+|E|)$ with sparse
aggregation. The context design matrix has dimension $3d+2$; the evidence
matrix adds seven columns. The finite gate evaluates five vectors over the
validation set, $O(5|\mathcal{V}_s|)$, and has no trainable parameters. Memory
is dominated by the graph and feature matrices. The method avoids dense
$n\times n$ adjacency construction.

\section{Experimental Design}
\label{sec:experiments}

\subsection{Research questions}
\label{sec:research-questions}

The evaluation is organized around five questions that separate representation
quality, evidence utility, selection behavior, external comparison, and
operational performance:
\begin{enumerate}[label=\textbf{RQ\arabic*.}]
  \item How much does label-free graph context improve over raw-feature and
  one-hop structural baselines?
  \item When does training-label-only relational evidence improve the
  label-free expert?
  \item Does the reduced adaptive gate improve the stability of the larger
  exploratory gate?
  \item How does PREF-Gate compare with established and recent external
  graph-fraud baselines under identical split memberships?
  \item Are ranking gains retained at a fixed 5\% review budget, and what
  calibration trade-offs remain?
\end{enumerate}

RQ1 is tested through the feature and graph-context controls; RQ2 and RQ3 use
the expert, mixture, and gate diagnostics; RQ4 uses the same-split external
comparisons and paired tests; and RQ5 uses review-budget and calibration
metrics. This mapping is fixed before the result interpretation.

\subsection{Datasets and splits}

\begin{table}[H]
\centering
\caption{Dataset characteristics and common evaluation protocol. Edge counts follow
deterministic undirected conversion, duplicate removal, and insertion of one
self-loop per node.}
\label{tab:datasets}
\small
\renewcommand{\arraystretch}{1.12}
\begin{tabular*}{\textwidth}{@{\extracolsep{\fill}}lrrrrr@{}}
\toprule
Dataset & Nodes & Edges & Features & Fraud nodes & Fraud rate \\
\midrule
Amazon & 11,944 & 8,847,096 & 25 & 821 & 6.87\% \\
YelpChi & 45,954 & 7,739,912 & 32 & 6,677 & 14.53\% \\
TFinance & 39,357 & 42,484,443 & 10 & 1,804 & 4.58\% \\
\bottomrule
\end{tabular*}
\vspace{0.35em}

\footnotesize\raggedright Each of five outer seeds uses a stratified 40\%/20\%/40\%
train/validation/test split. The same split indices are used by every method.
\end{table}

Amazon and YelpChi are public opinion-fraud graphs distributed through the DGL
fraud dataset interface, with original benchmark provenance
\cite{rayana2015yelpchi,mccauley2013amazon}. TFinance is a public transaction
fraud graph. The three datasets differ substantially in feature dimension,
graph density, and positive rate, which is important for testing whether a
single evidence policy is uniformly appropriate.

Five outer seeds, 11, 23, 37, 53, and 71, generate stratified
40\%/20\%/40\% splits. The serialized indices and their SHA256 hashes are
fixed. Every method uses exactly the same train, validation, and test
memberships for a given dataset and seed.

\subsection{Comparison roster}

The 14-method roster contains:
\begin{itemize}
  \item \textbf{Feature MLP}: standardized raw features and a one-hidden-layer
  multilayer perceptron;
  \item \textbf{One-hop mean-aggregation MLP}: raw and one-hop mean features with the same
  MLP training rule;
  \item \textbf{BPM structural}: an independently seeded structural expert;
  \item \textbf{Neighbor risk}: the smoothed training-neighbor risk;
  \item \textbf{EB neighbor evidence}: the empirical-Bayes posterior mean;
  \item \textbf{Label-free context} and \textbf{evidence context}: the two
  final PREF-Gate experts;
  \item \textbf{Exploratory full gate}: the original 35-candidate selector;
  \item \textbf{CARE-GNN}, \textbf{PC-GNN}, and \textbf{BWGNN}: established
  fraud/anomaly GNN baselines;
  \item \textbf{PMP} and \textbf{ConsisGAD}: recent ICLR 2024 baselines; and
  \item \textbf{PREF-Gate adaptive gate}: the final five-candidate method.
\end{itemize}

External adapters preserve the official model core while replacing internal
split generation with caller-supplied frozen indices. Checkpoints are selected
on validation data, non-training labels are masked where the original
transductive method consumes unlabeled nodes, and test inference is deferred.
The stock loops are not used when they evaluate test data every epoch.

\subsection{Metrics}

The primary ranking metric is AUPRC. Precision@5\% and Recall@5\% measure
performance under a fixed review budget. Macro-F1 uses a threshold selected on
validation. AUROC is secondary because it can appear optimistic under strong
imbalance. Brier score and equal-width 10-bin ECE assess probability quality;
lower values are better.

\subsection{Statistical policy}

All statistical tests operate on paired outer splits, never on pooled nodes.
We report two-sided paired $t$-tests, Wilcoxon signed-rank tests, win/tie/loss
counts, and paired Cohen $d_z$. Five splits provide limited non-parametric
resolution: when all five differences are non-zero and have the same sign, the
minimum exact two-sided Wilcoxon $p$ is 0.0625. Accordingly, effect sizes and
cross-split consistency are interpreted alongside $p$-values. The complete
CSV also reports a conservative Holm adjustment across the generated test
family.

\subsection{Execution and failure accounting}

\begin{table}[H]
\centering
\caption{Execution status by outcome category.}
\label{tab:execution}
\small
\renewcommand{\arraystretch}{1.12}
\resizebox{\textwidth}{!}{%
\begin{tabular}{lrrrr}
\toprule
Block & Planned & Successful & Failed method & Failed infrastructure \\
\midrule
Locked 13-method matrix & 195 & 185 & 5 & 5 \\
Adaptive-gate confirmation & 15 & 15 & 0 & 0 \\
\bottomrule
\end{tabular}}

\footnotesize The five method failures are PC-GNN on TFinance. The five
infrastructure failures are PMP on TFinance, where all attempts exceeded the
available host-memory envelope before producing valid metrics.
\end{table}

The original Cartesian matrix contains $3\times5\times13=195$ rows. Of these,
185 are successful. PC-GNN on TFinance is documented as a method-compatibility
failure, and PMP on TFinance is documented as an infrastructure failure after
repeated CPU and CUDA-DGL attempts exceeded the 14~GiB host-memory envelope.
No number is copied, estimated, or imputed for either failure block. The
adaptive gate adds 15 successful runs, producing 200 valid metric rows across
the 210 planned method--dataset--split combinations.

\section{Results}
\label{sec:results}

\subsection{Overall comparison}

\begin{table}[H]
\centering
\caption{Main same-protocol results (mean $\pm$ sample standard deviation over five outer splits). A dash denotes a documented method or infrastructure failure rather than an imputed value. Highest means among successfully completed methods in the same-protocol evaluation are bold; exact ties are highlighted equally.}
\label{tab:main-results}
\renewcommand{\arraystretch}{1.08}
\resizebox{\textwidth}{!}{%
\begin{tabular}{llcccc}
\toprule
Dataset & Method & Precision@5\% & Recall@5\% & AUPRC & Macro-F1 \\
\midrule
Amazon & Feature MLP & 0.9481 $\pm$ 0.0121 & 0.6888 $\pm$ 0.0088 & 0.8845 $\pm$ 0.0087 & 0.9197 $\pm$ 0.0060 \\
 & One-hop mean-aggregation MLP & 0.9456 $\pm$ 0.0118 & 0.6869 $\pm$ 0.0086 & 0.8761 $\pm$ 0.0141 & 0.9065 $\pm$ 0.0078 \\
 & BPM structural & 0.9473 $\pm$ 0.0121 & 0.6881 $\pm$ 0.0088 & 0.8765 $\pm$ 0.0127 & 0.9057 $\pm$ 0.0050 \\
 & Label-free context & \textbf{0.9590 $\pm$ 0.0108} & \textbf{0.6967 $\pm$ 0.0079} & \textbf{0.9085 $\pm$ 0.0080} & \textbf{0.9227 $\pm$ 0.0059} \\
 & Evidence context & 0.8469 $\pm$ 0.0317 & 0.6152 $\pm$ 0.0230 & 0.7715 $\pm$ 0.0157 & 0.8458 $\pm$ 0.0132 \\
 & Exploratory full gate & 0.9582 $\pm$ 0.0089 & 0.6960 $\pm$ 0.0064 & 0.9050 $\pm$ 0.0088 & 0.9214 $\pm$ 0.0049 \\
 & CARE-GNN & 0.9414 $\pm$ 0.0107 & 0.6839 $\pm$ 0.0077 & 0.8422 $\pm$ 0.0130 & 0.9099 $\pm$ 0.0064 \\
 & PC-GNN & 0.8929 $\pm$ 0.0322 & 0.6486 $\pm$ 0.0234 & 0.8097 $\pm$ 0.0303 & 0.8762 $\pm$ 0.0206 \\
 & BWGNN & 0.9515 $\pm$ 0.0048 & 0.6912 $\pm$ 0.0035 & 0.8601 $\pm$ 0.0084 & 0.9142 $\pm$ 0.0025 \\
 & PMP & 0.9523 $\pm$ 0.0096 & 0.6918 $\pm$ 0.0070 & 0.8690 $\pm$ 0.0124 & 0.9129 $\pm$ 0.0069 \\
 & ConsisGAD & 0.9515 $\pm$ 0.0138 & 0.6912 $\pm$ 0.0100 & 0.8696 $\pm$ 0.0186 & 0.9180 $\pm$ 0.0039 \\
\rowcolor{prefshade} & \textbf{PREF-Gate (adaptive gate)} & \textbf{0.9590 $\pm$ 0.0108} & \textbf{0.6967 $\pm$ 0.0079} & \textbf{0.9085 $\pm$ 0.0080} & \textbf{0.9227 $\pm$ 0.0059} \\
\midrule
YelpChi & Feature MLP & 0.6933 $\pm$ 0.0120 & 0.2388 $\pm$ 0.0041 & 0.5245 $\pm$ 0.0091 & 0.7105 $\pm$ 0.0058 \\
 & One-hop mean-aggregation MLP & 0.8054 $\pm$ 0.0144 & 0.2774 $\pm$ 0.0050 & 0.6136 $\pm$ 0.0072 & 0.7501 $\pm$ 0.0041 \\
 & BPM structural & 0.8083 $\pm$ 0.0077 & 0.2784 $\pm$ 0.0027 & 0.6131 $\pm$ 0.0072 & 0.7487 $\pm$ 0.0029 \\
 & Label-free context & \textbf{0.9567 $\pm$ 0.0040} & \textbf{0.3295 $\pm$ 0.0014} & \textbf{0.8104 $\pm$ 0.0043} & \textbf{0.8405 $\pm$ 0.0032} \\
 & Evidence context & 0.7778 $\pm$ 0.0169 & 0.2679 $\pm$ 0.0058 & 0.5932 $\pm$ 0.0071 & 0.7323 $\pm$ 0.0021 \\
 & Exploratory full gate & 0.9530 $\pm$ 0.0108 & 0.3283 $\pm$ 0.0037 & 0.8090 $\pm$ 0.0062 & 0.8396 $\pm$ 0.0025 \\
 & CARE-GNN & 0.5250 $\pm$ 0.0094 & 0.1808 $\pm$ 0.0032 & 0.3813 $\pm$ 0.0022 & 0.6447 $\pm$ 0.0024 \\
 & PC-GNN & 0.6707 $\pm$ 0.0160 & 0.2310 $\pm$ 0.0055 & 0.4925 $\pm$ 0.0048 & 0.6998 $\pm$ 0.0029 \\
 & BWGNN & 0.6857 $\pm$ 0.0077 & 0.2362 $\pm$ 0.0027 & 0.5105 $\pm$ 0.0095 & 0.7017 $\pm$ 0.0063 \\
 & PMP & 0.8980 $\pm$ 0.0233 & 0.3093 $\pm$ 0.0080 & 0.7340 $\pm$ 0.0179 & 0.8012 $\pm$ 0.0080 \\
 & ConsisGAD & 0.6800 $\pm$ 0.0195 & 0.2342 $\pm$ 0.0067 & 0.5397 $\pm$ 0.0140 & 0.7198 $\pm$ 0.0074 \\
\rowcolor{prefshade} & \textbf{PREF-Gate (adaptive gate)} & \textbf{0.9567 $\pm$ 0.0040} & \textbf{0.3295 $\pm$ 0.0014} & \textbf{0.8104 $\pm$ 0.0043} & \textbf{0.8405 $\pm$ 0.0032} \\
\midrule
TFinance & Feature MLP & 0.6266 $\pm$ 0.0118 & 0.6849 $\pm$ 0.0129 & 0.7211 $\pm$ 0.0158 & 0.8352 $\pm$ 0.0083 \\
 & One-hop mean-aggregation MLP & 0.7046 $\pm$ 0.0096 & 0.7700 $\pm$ 0.0105 & 0.8101 $\pm$ 0.0113 & 0.8842 $\pm$ 0.0083 \\
 & BPM structural & 0.7048 $\pm$ 0.0078 & 0.7703 $\pm$ 0.0085 & 0.8075 $\pm$ 0.0137 & 0.8839 $\pm$ 0.0070 \\
 & Label-free context & 0.7777 $\pm$ 0.0111 & 0.8499 $\pm$ 0.0121 & 0.8860 $\pm$ 0.0090 & 0.9204 $\pm$ 0.0052 \\
 & Evidence context & 0.7332 $\pm$ 0.0154 & 0.8014 $\pm$ 0.0169 & 0.8350 $\pm$ 0.0071 & 0.8830 $\pm$ 0.0037 \\
 & Exploratory full gate & \textbf{0.7799 $\pm$ 0.0125} & \textbf{0.8524 $\pm$ 0.0136} & \textbf{0.8913 $\pm$ 0.0106} & \textbf{0.9212 $\pm$ 0.0061} \\
 & CARE-GNN & 0.6155 $\pm$ 0.0087 & 0.6727 $\pm$ 0.0095 & 0.7259 $\pm$ 0.0110 & 0.8598 $\pm$ 0.0076 \\
 & PC-GNN & \textit{N/A} & \textit{N/A} & \textit{N/A} & \textit{N/A} \\
 & BWGNN & 0.6764 $\pm$ 0.1327 & 0.7393 $\pm$ 0.1451 & 0.7648 $\pm$ 0.1922 & 0.8538 $\pm$ 0.0790 \\
 & PMP & \textit{N/A} & \textit{N/A} & \textit{N/A} & \textit{N/A} \\
 & ConsisGAD & 0.7574 $\pm$ 0.0091 & 0.8277 $\pm$ 0.0100 & 0.8668 $\pm$ 0.0090 & 0.9057 $\pm$ 0.0058 \\
\rowcolor{prefshade} & \textbf{PREF-Gate (adaptive gate)} & \textbf{0.7799 $\pm$ 0.0125} & \textbf{0.8524 $\pm$ 0.0136} & \textbf{0.8913 $\pm$ 0.0106} & \textbf{0.9212 $\pm$ 0.0061} \\
\bottomrule
\end{tabular}}
\end{table}

Table~\ref{tab:main-results} gives the central result. PREF-Gate attains the
highest mean values on all reported primary measures among successfully
completed methods under the frozen same-protocol evaluation. On Amazon it obtains AUPRC
$0.9085\pm0.0080$, Precision@5\% $0.9590\pm0.0108$, and Macro-F1
$0.9227\pm0.0059$. On YelpChi it obtains AUPRC $0.8104\pm0.0043$ and
Macro-F1 $0.8405\pm0.0032$. On TFinance it reaches AUPRC
$0.8913\pm0.0106$ and Macro-F1 $0.9212\pm0.0061$.

\begin{figure}[H]
  \centering
  \includegraphics[width=0.88\linewidth]{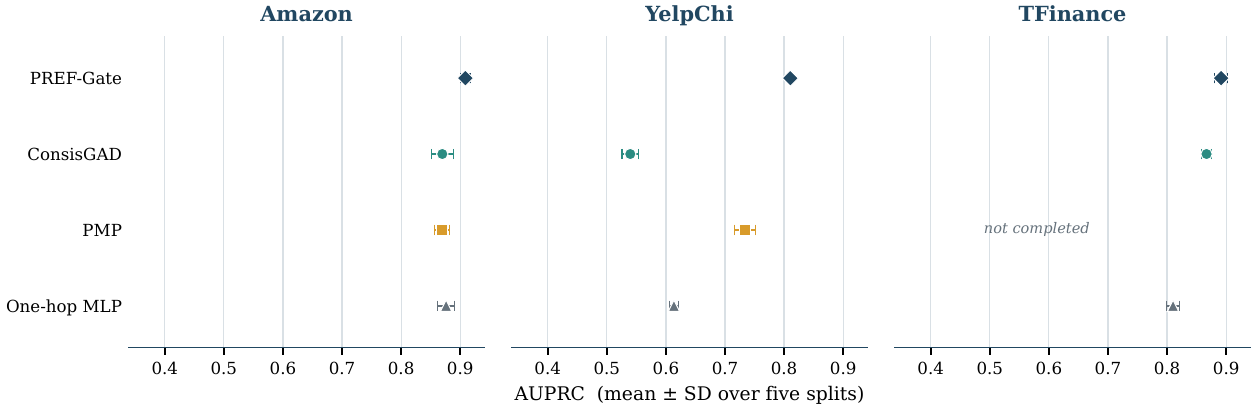}
  \caption{AUPRC point estimates with one sample-standard-deviation horizontal
  error bar. The missing PMP--TFinance result is marked as not completed rather
  than imputed.}
  \label{fig:main-auprc}
\end{figure}

The adaptive method ties the label-free expert on Amazon and YelpChi because
the gate selects that expert in every split. It ties the exploratory full gate
on TFinance because both select the same evidence mixtures. These exact ties
are expected consequences of a validation-routed framework, not duplicated
independent estimates.

\subsection{RQ1: value of label-free graph context}

The label-free expert improves substantially over the raw-feature and
one-hop mean-aggregation MLP baselines. Relative to the one-hop
mean-aggregation MLP, its AUPRC gain is 0.0324 on Amazon, 0.1968 on YelpChi,
and 0.0759 on TFinance. The
largest gain occurs on YelpChi, where absolute node-to-neighborhood residuals
and tree-based interactions appear especially informative. Macro-F1 follows
the same pattern.

This result clarifies the source of performance. It is not sufficient to
attribute every gain to training-label risk. Much of PREF-Gate's strength comes
from a label-free graph representation that exposes raw context and local
disagreement to a flexible classifier. The method remains graph based because
two thirds of its main feature block depend on one-hop graph neighborhoods,
but it does not require a deep message-passing optimizer.

\subsection{RQ2: when evidence helps}

\begin{table}[H]
\centering
\caption{Mechanism comparison. Differences are PREF-Gate minus the comparator.}
\label{tab:ablation}
\small
\renewcommand{\arraystretch}{1.10}
\resizebox{\linewidth}{!}{%
\begin{tabular}{llrrr}
\toprule
Dataset & Comparator & $\Delta$AUPRC & $\Delta$Macro-F1 & Gate interpretation \\
\midrule
Amazon & Label-free context & +0.0000 & +0.0000 & fallback selected \\
Amazon & Evidence context & +0.1370 & +0.0769 & gate avoids unsupported full evidence \\
Amazon & Exploratory full gate & +0.0035 & +0.0013 & reduced candidate variance \\
YelpChi & Label-free context & +0.0000 & +0.0000 & fallback selected \\
YelpChi & Evidence context & +0.2173 & +0.1082 & gate avoids unsupported full evidence \\
YelpChi & Exploratory full gate & +0.0014 & +0.0009 & reduced candidate variance \\
\rowcolor{prefshade} TFinance & Label-free context & +0.0053 & +0.0008 & evidence accepted \\
TFinance & Evidence context & +0.0563 & +0.0382 & gate avoids unsupported full evidence \\
TFinance & Exploratory full gate & +0.0000 & +0.0000 & reduced candidate variance \\
\bottomrule
\end{tabular}}
\end{table}

\begin{table}[H]
\centering
\caption{Validation-only adaptive-gate decisions across five splits.}
\label{tab:gate-decisions}
\small
\renewcommand{\arraystretch}{1.12}
\begin{tabular}{lp{0.58\linewidth}r}
\toprule
Dataset & Selected candidate & Splits \\
\midrule
Amazon & Context expert ($C$) & 5/5 \\
YelpChi & Context expert ($C$) & 5/5 \\
\rowcolor{prefshade} TFinance & $0.75C+0.25E$ & 4/5 \\
\rowcolor{prefshade} TFinance & $0.25C+0.75E$ & 1/5 \\
\bottomrule
\end{tabular}
\end{table}

The stand-alone evidence context expert is weaker than the label-free expert
on every dataset. This negative result is important: concatenating
training-label evidence to a strong representation is not automatically
beneficial. The finite gate instead asks whether a small amount of evidence
improves validation.

On Amazon and YelpChi, label-free context is selected in 10 of 10 runs.
Evidence mixtures have lower validation scores and are rejected. On TFinance,
four splits select $0.75p^{(C)}+0.25p^{(E)}$, while one selects
$0.25p^{(C)}+0.75p^{(E)}$. The resulting AUPRC exceeds label-free context by
0.0053, with wins on all five paired splits. The paired $t$-test is
$p=0.0030$, while the exact Wilcoxon test is $p=0.0625$ because $n=5$.

\begin{figure}[H]
  \centering
  \includegraphics[width=0.82\linewidth]{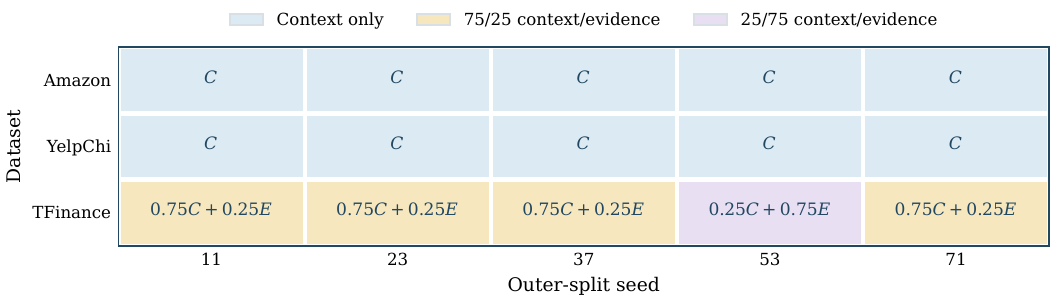}
  \caption{Validation-only candidate selected for each outer split. The matrix
  exposes uniform context fallback on Amazon and YelpChi and evidence mixtures
  on TFinance.}
  \label{fig:gate}
\end{figure}

These decisions support a bounded mechanism interpretation. Evidence helps on
TFinance under the evaluated protocol, but the framework's robustness on
Amazon and YelpChi comes from rejecting it. The paper therefore treats
adaptive use, rather than universal use, as the contribution.

\subsection{RQ3: stability relative to the exploratory gate}

The 35-candidate gate obtains AUPRC 0.9050 on Amazon and 0.8090 on YelpChi.
The reduced gate improves those values by 0.0035 and 0.0014, respectively, by
avoiding weak structural or neighbor-risk mixtures selected for tiny
validation gains. On TFinance the two gates agree and both obtain 0.8913.

The improvement is not statistically decisive with five splits: the Amazon
paired $t$-test is $p=0.481$ and the YelpChi test is $p=0.374$. The value of the
revision is instead methodological. It reduces the candidate count from 35 to
5, aligns selection with the evidence question, and removes avoidable
multiple-comparison variance without reducing the highest dataset-level
performance observed within the present evaluation.

\subsection{RQ4: external baselines}

\begin{table}[H]
\centering
\caption{External comparison under identical split memberships. Positive differences favor PREF-Gate.}
\label{tab:external}
\small
\renewcommand{\arraystretch}{1.10}
\resizebox{\textwidth}{!}{%
\begin{tabular}{llrrrr}
\toprule
Dataset & External method & External AUPRC & PREF-Gate AUPRC & Difference & AUPRC W/T/L \\
\midrule
Amazon & CARE-GNN & 0.8422 & 0.9085 & +0.0663 & 5/0/0 \\
Amazon & PC-GNN & 0.8097 & 0.9085 & +0.0987 & 5/0/0 \\
Amazon & BWGNN & 0.8601 & 0.9085 & +0.0484 & 5/0/0 \\
Amazon & PMP & 0.8690 & 0.9085 & +0.0395 & 5/0/0 \\
Amazon & ConsisGAD & 0.8696 & 0.9085 & +0.0389 & 5/0/0 \\
\addlinespace[2pt]
YelpChi & CARE-GNN & 0.3813 & 0.8104 & +0.4291 & 5/0/0 \\
YelpChi & PC-GNN & 0.4925 & 0.8104 & +0.3180 & 5/0/0 \\
YelpChi & BWGNN & 0.5105 & 0.8104 & +0.2999 & 5/0/0 \\
YelpChi & PMP & 0.7340 & 0.8104 & +0.0764 & 5/0/0 \\
YelpChi & ConsisGAD & 0.5397 & 0.8104 & +0.2707 & 5/0/0 \\
\addlinespace[2pt]
TFinance & CARE-GNN & 0.7259 & 0.8913 & +0.1654 & 5/0/0 \\
TFinance & PC-GNN & \textit{N/A} & 0.8913 & \textit{N/A} & \textit{N/A} \\
TFinance & BWGNN & 0.7648 & 0.8913 & +0.1265 & 5/0/0 \\
TFinance & PMP & \textit{N/A} & 0.8913 & \textit{N/A} & \textit{N/A} \\
TFinance & ConsisGAD & 0.8668 & 0.8913 & +0.0245 & 5/0/0 \\
\bottomrule
\end{tabular}}
\end{table}

PREF-Gate has positive mean-AUPRC margins over every completed external baseline
in the same-protocol evaluation. The highest
completed external scores are ConsisGAD on Amazon (0.8696), PMP on YelpChi
(0.7340), and ConsisGAD on TFinance (0.8668); corresponding PREF-Gate margins are
0.0389, 0.0764, and 0.0245, with five paired-split wins in each comparison.
PC-GNN and PMP did not complete TFinance under the frozen environment and are
not used to support a ranking beyond the completed comparison roster.

CARE-GNN and PC-GNN remain important established comparators. PREF-Gate improves
over CARE-GNN by 0.0663 AUPRC on Amazon, 0.4291 on YelpChi, and 0.1654 on
TFinance. It improves over PC-GNN by 0.0987 on Amazon and 0.3180 on YelpChi;
PC-GNN's TFinance configuration is not reported because the official method
semantics could not be preserved by a valid adapter on that graph.

\begin{table}[H]
\centering
\caption{Paired split-level AUPRC tests. Tests are two-sided. With five non-zero pairs, the minimum attainable exact two-sided Wilcoxon $p$ is 0.0625.}
\label{tab:statistics}
\small
\renewcommand{\arraystretch}{1.08}
\resizebox{\textwidth}{!}{%
\begin{tabular}{llrrrrr}
\toprule
Dataset & Comparator & Mean difference & W/T/L & Paired $t$ $p$ & Wilcoxon $p$ & Cohen $d_z$ \\
\midrule
Amazon & One-hop mean-aggregation MLP & +0.0324 & 5/0/0 & 0.002652 & 0.0625 & +2.975 \\
Amazon & CARE-GNN & +0.0663 & 5/0/0 & 0.0008361 & 0.0625 & +4.035 \\
Amazon & PC-GNN & +0.0987 & 5/0/0 & 0.002482 & 0.0625 & +3.028 \\
Amazon & BWGNN & +0.0484 & 5/0/0 & 0.002035 & 0.0625 & +3.193 \\
Amazon & PMP & +0.0395 & 5/0/0 & 0.003534 & 0.0625 & +2.753 \\
Amazon & ConsisGAD & +0.0389 & 5/0/0 & 0.00293 & 0.0625 & +2.896 \\
\addlinespace[2pt]
YelpChi & One-hop mean-aggregation MLP & +0.1968 & 5/0/0 & 3.498e-07 & 0.0625 & +28.768 \\
YelpChi & CARE-GNN & +0.4291 & 5/0/0 & 3.797e-09 & 0.0625 & +89.162 \\
YelpChi & PC-GNN & +0.3180 & 5/0/0 & 1.543e-09 & 0.0625 & +111.682 \\
YelpChi & BWGNN & +0.2999 & 5/0/0 & 5.889e-07 & 0.0625 & +25.253 \\
YelpChi & PMP & +0.0764 & 5/0/0 & 0.001276 & 0.0625 & +3.613 \\
YelpChi & ConsisGAD & +0.2707 & 5/0/0 & 1.313e-06 & 0.0625 & +20.661 \\
\addlinespace[2pt]
TFinance & One-hop mean-aggregation MLP & +0.0812 & 5/0/0 & 3.922e-06 & 0.0625 & +15.707 \\
TFinance & CARE-GNN & +0.1654 & 5/0/0 & 6.682e-06 & 0.0625 & +13.743 \\
TFinance & BWGNN & +0.1265 & 5/0/0 & 0.2159 & 0.0625 & +0.657 \\
TFinance & ConsisGAD & +0.0245 & 5/0/0 & 0.0006315 & 0.0625 & +4.339 \\
\bottomrule
\end{tabular}}
\end{table}

The paired $t$-tests are small for many comparisons, particularly on YelpChi.
Nevertheless, exact Wilcoxon values are generally 0.0625 for five consistent
wins, and broad Holm correction is conservative. The result should therefore
be read as a reproducible, same-protocol ranking with consistent effect
direction, not as high-powered evidence over a population of datasets.

\subsection{Protocol-specific comparison statement}

Within the present frozen evaluation design, PREF-Gate attains the highest mean
AUPRC among completed methods and has positive margins against
every completed external baseline. This statement is restricted to the
completed evaluation roster. Recent methods that could not satisfy the common
dataset, split, label-boundary, dependency, and deferred-test rules are not
assigned quantitative outcomes.

The positive external margins support leading performance within the completed
same-protocol comparison. They do not establish a ranking beyond methods
evaluated with the same supervision ratios and test policy.

\subsection{RQ5: constrained review budget}

\begin{figure}[H]
  \centering
  \includegraphics[width=0.96\linewidth]{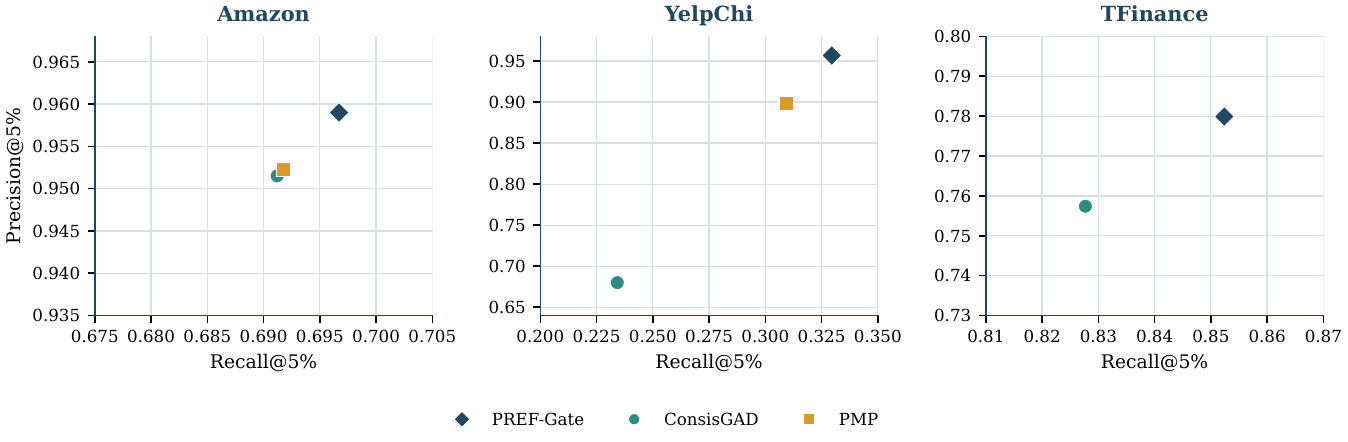}
  \caption{Precision--recall operating points at a 5\% review budget. Higher
  values on both axes are preferable; missing PMP--TFinance results are not
  plotted.}
  \label{fig:budget}
\end{figure}

Within the completed frozen-protocol comparison, PREF-Gate has the highest mean
Precision@5\% and Recall@5\%. Precision@5\% is 0.9590 on Amazon, 0.9567 on
YelpChi, and 0.7799 on TFinance. Recall@5\% is 0.6967, 0.3295, and 0.8524,
respectively. The apparent disparity between YelpChi precision and recall
reflects the dataset's 14.53\% fraud prevalence: a queue containing only 5\%
of test nodes cannot recover all positives even when it is highly pure.

Compared with the highest-AUPRC recent external baseline that successfully
completed each dataset under the frozen protocol, the adaptive gate increases
Precision@5\% by 0.0075 over ConsisGAD on Amazon, 0.0587 over PMP on YelpChi,
and 0.0226 over ConsisGAD on TFinance. Thus, the same-protocol ranking
improvement is retained at an operationally constrained point.

\subsection{Calibration diagnostics}

\begin{table}[H]
\centering
\caption{Probability-quality diagnostics. Lower is better for Brier and ECE.}
\label{tab:calibration}
\small
\renewcommand{\arraystretch}{1.10}
\resizebox{\linewidth}{!}{%
\begin{tabular}{llcc}
\toprule
Dataset & Method & Brier & ECE (10 bins) \\
\midrule
\rowcolor{prefshade} Amazon & PREF-Gate (adaptive gate) & 0.0165 $\pm$ 0.0011 & 0.0110 $\pm$ 0.0016 \\
Amazon & Label-free context & 0.0165 $\pm$ 0.0011 & 0.0110 $\pm$ 0.0016 \\
Amazon & ConsisGAD & 0.0176 $\pm$ 0.0011 & 0.0095 $\pm$ 0.0016 \\
\midrule
\rowcolor{prefshade} YelpChi & PREF-Gate (adaptive gate) & 0.0753 $\pm$ 0.0007 & 0.1077 $\pm$ 0.0024 \\
YelpChi & Label-free context & 0.0753 $\pm$ 0.0007 & 0.1077 $\pm$ 0.0024 \\
YelpChi & ConsisGAD & 0.0955 $\pm$ 0.0029 & 0.0515 $\pm$ 0.0187 \\
\midrule
\rowcolor{prefshade} TFinance & PREF-Gate (adaptive gate) & 0.0182 $\pm$ 0.0022 & 0.0585 $\pm$ 0.0078 \\
TFinance & Label-free context & 0.0195 $\pm$ 0.0019 & 0.0524 $\pm$ 0.0063 \\
TFinance & ConsisGAD & 0.0136 $\pm$ 0.0009 & 0.0117 $\pm$ 0.0015 \\
\bottomrule
\end{tabular}}
\end{table}

\begin{figure}[H]
  \centering
  \includegraphics[width=0.96\linewidth]{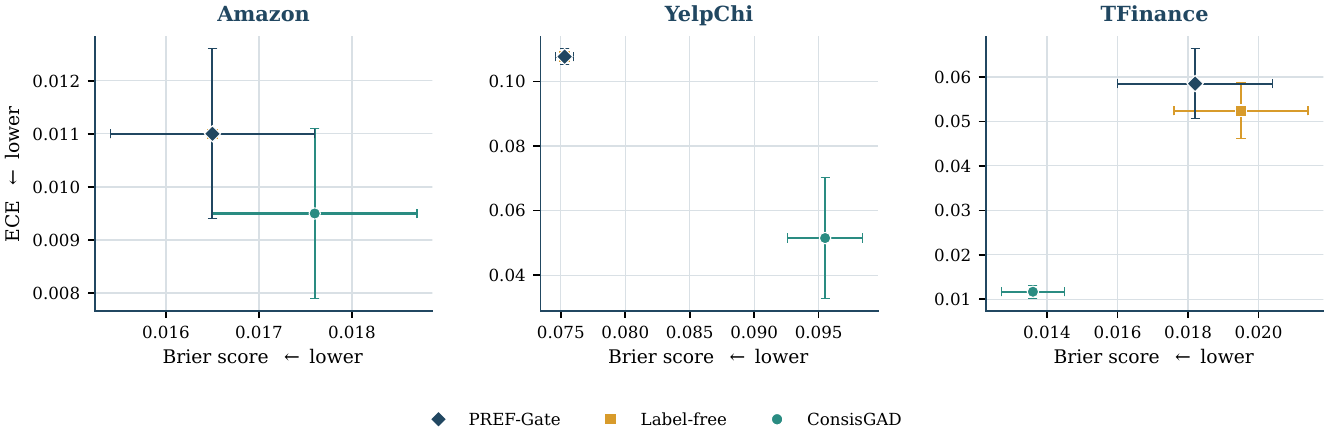}
  \caption{Joint Brier--ECE calibration diagnostics with one-standard-deviation
  error bars. Lower values on both axes are preferable; exact overlap denotes
  identical predictions.}
  \label{fig:calibration}
\end{figure}

Calibration is mixed. PREF-Gate's Brier score is low on Amazon (0.0165) and
TFinance (0.0182), but its ECE is 0.1077 on YelpChi and 0.0585 on TFinance.
The adaptive gate selects candidates for discrimination, not calibration; no
post-hoc calibrator is fitted. The near identity between PREF-Gate and the
label-free expert on Amazon and YelpChi is again expected because the gate
chooses that candidate.

These results motivate separate deployment calibration. A platform could fit
a validation-only calibrator after the ranking model is fixed, but doing so
would constitute an additional method component and is not retroactively added
to this benchmark. The current paper reports the probabilities actually
produced by the evaluated method.

\section{Discussion}
\label{sec:discussion}

\subsection{Why the simple context expert is strong}

Fraud graphs often contain informative node attributes. The one-hop mean and
absolute residual transform these attributes into local comparison features:
a node can be suspicious because its value is extreme globally, because its
neighborhood is unusual, or because it disagrees with that neighborhood.
Tree boosting can partition these interactions without forcing the same
linear message transformation across all nodes.

The result does not imply that learned GNNs are unnecessary in general. It
shows that a transparent graph-context representation is a strong comparator
on these three benchmarks and supervision ratios. A fair graph-fraud study
should include such a baseline rather than attributing all graph gains to a
neural architecture.

\subsection{Why evidence is dataset dependent}

Evidence can fail for three reasons. First, eligible support may be sparse, so
local risk mostly reflects the global prior. Second, fraud may be
heterophilically connected, making neighbor labels misleading. Third, the
label-free representation may already encode the discriminative signal, while
seven evidence columns add variance. Posterior variance and displacement
describe these conditions but cannot create information that is absent.

TFinance exhibits the opposite pattern: small evidence mixtures improve
AUPRC consistently across all five splits. Its low fraud prevalence and dense
normalized graph make support-aware neighborhood risk complementary to the
label-free context. The dominant 75/25 mixture indicates that evidence should
modify, not replace, the stronger context probability.

\subsection{What the fallback means}

A reviewer may reasonably ask whether a method that falls back to an internal
expert is still a single method. The answer depends on whether the routing rule
is specified before test evaluation. PREF-Gate fixes the expert definitions,
weights, score, tie-breaking, and threshold rule. The selected candidate is an
output of the learning protocol in the same sense that a validation-selected
checkpoint or regularization value is an output.

The fallback should not be hidden as a universal evidence-fusion gain. It is
reported explicitly in Table~\ref{tab:gate-decisions}. On Amazon and YelpChi,
the final prediction equals the label-free expert; on TFinance it equals a
selected mixture. This transparency is central to the claim.

\subsection{Implications for knowledge-based decision systems}

Knowledge-based decision systems require more than a high test score. They
need auditable information boundaries, explicit reliability descriptors, a
manageable decision space, and interpretable fallback behavior. PREF-Gate treats
neighborhood statistics as provenance-typed knowledge rather than ordinary
features. Its evidence vector records why a rate may be unreliable, and its
finite candidate set prevents an unconstrained ensemble search after outcomes
have been observed.

For an operational workflow, the selected model should be accompanied by the
training split hash, graph snapshot, evidence concentration, candidate score
trace, threshold, and review-budget policy. These artifacts permit later
audits under label delay or graph drift.

\subsection{Recent-baseline coverage}

Recency alone is not a fairness criterion. PMP and ConsisGAD were included
because their implementations could be adapted to all required label and split
boundaries. DGA-GNN and DiG-In-GNN were screened, but their local stock paths
required unavailable stacks or internal split/test behavior that had not
passed the deferred-test adapter gate. Several 2025 methods did not provide the
same supervised three-dataset setting. Reporting their paper numbers beside
the present results would mix protocols and weaken, rather than strengthen,
the evidence.

\section{Threats to Validity and Limitations}
\label{sec:limitations}

\paragraph{Post-development protocol revision}
The adaptive gate was motivated by a completed exploratory benchmark that
revealed selection variance in a 35-candidate gate. The revised candidate
family and validation rule were frozen before the 15 adaptive-gate executions,
and old external baseline rows were not rerun or overwritten. Nevertheless,
the revision is part of the same research program rather than an independent
external replication. Future work should test the fixed five-candidate method
on new datasets and temporal splits.

\paragraph{Dataset scope}
Only three public graphs are used. Five random outer splits measure sensitivity
to node membership but are not five independent datasets. Results may differ
under event-time splits, inductive arrival, or industrial graphs with delayed
labels.

\paragraph{Baseline availability}
PC-GNN and PMP do not have valid TFinance metrics in the present environment.
These failures are visible, but they reduce the completeness of the TFinance
roster. ConsisGAD provides a successful recent comparison on all three
datasets. No claim is made against excluded 2025 methods.

\paragraph{One-hop aggregation baseline}
The one-hop mean-aggregation MLP baseline concatenates raw attributes and a
materialized one-hop neighbor mean, then fits an MLP. It is not presented as
GraphSAGE because it does not train an end-to-end neighborhood-sampling
encoder. The original GraphSAGE paper remains cited only as background for the
mean-aggregation operator. External fraud baselines use their audited model
cores.

\paragraph{Framework nomenclature}
The selected predictor is described throughout as PREF-Gate, a hybrid graph
expert framework. The title, abstract, method section, and cover letter state
that boundary. The paper does not call the method a GNN or attribute
tree-ensemble gains to an unexecuted neural architecture.

\paragraph{Calibration}
The gate optimizes a discrimination objective. ECE remains elevated on YelpChi
and TFinance, so scores should not be interpreted as deployment-ready
probabilities without validation-only calibration and drift monitoring.

\paragraph{Computational environment}
The host provides two V100S GPUs but only 14~GiB of system memory. PMP's
TFinance preprocessing exceeds that host-memory envelope even with small
batches; GPU memory does not repair a CPU-side graph materialization failure.
This is recorded as infrastructure failure rather than method inferiority.

\paragraph{No causal explanation}
The evidence descriptors are auditable statistics, but the tree ensembles do
not provide a causal explanation for a prediction. Feature importance or
post-hoc attribution could be added, but it would not establish causality.

\section{Reproducibility}
\label{sec:reproducibility}

Every successful row records dataset, outer seed, method, selected candidate,
threshold, all metrics, split-file hash, train/validation/test index hashes,
protocol hash, method-config hash, source hash, elapsed time, prediction-file
hash, and completion time. Candidate diagnostics record all five validation
scores and a post-freeze oracle analysis that is never fed back into selection.

The author-verified reproducibility record contains the combined long-form
metrics, aggregate results, paired tests, status matrices, frozen protocol,
method configuration, source hashes, and SHA256 manifest. Code, split
manifests, and processed evaluation artifacts will be released through an
author-verified repository upon acceptance. Executing the runner a second time
finds no \texttt{NOT-RUN} row and therefore performs no duplicate experiment.

\section{Conclusion}
\label{sec:conclusion}

PREF-Gate addresses graph fraud detection as a provenance-constrained evidence
decision problem. It combines a label-free neighborhood-context expert with a
self-excluded, training-label-only empirical-Bayes evidence expert. A finite
five-candidate gate selects either expert or a fixed probability mixture using
validation data and fixes the threshold before test inference.

Across Amazon, YelpChi, and TFinance, PREF-Gate obtains AUPRC 0.9085, 0.8104, and
0.8913. These are the highest mean AUPRC values among successfully completed
methods in the same-protocol evaluation, with margins of 0.0389,
0.0764, and 0.0245 over the highest-scoring completed external baseline on
each dataset. The gate rejects evidence on Amazon and YelpChi and accepts it
on TFinance. The central finding is therefore not that relational evidence
always helps, but that a leakage-safe validation policy can use it without
forcing it.

The contribution is the coupling of typed relational evidence, an enforceable
label-access contract, reliability-aware evidence descriptors, and a finite
decision policy. Its empirical gains are limited to the completed
same-protocol evaluation and should not be interpreted as a ranking across
methods or data regimes beyond the present design. Testing the fixed policy on
temporal and inductive graphs is the most important next step.

\clearpage
\appendix
\section{Complete Aggregate Benchmark}
\label{app:complete}

The main text emphasizes primary operational metrics and the methods most
relevant to the final claim. Table~\ref{tab:all-summary} lists every successful
method--dataset aggregate, including the stand-alone neighbor-risk and
empirical-Bayes evidence outputs. These stand-alone probabilities are useful
mechanism controls but are not intended as final classifiers.

\begin{landscape}
\scriptsize
\renewcommand{\arraystretch}{1.04}
\begin{longtable}{llrrrrrr}
\caption{Complete aggregate AUPRC, Macro-F1, and AUROC results.}\label{tab:all-summary}\\
\toprule
Dataset & Method & $n$ & AUPRC & SD & Macro-F1 & AUROC & Status note \\
\midrule
\endfirsthead
\toprule
Dataset & Method & $n$ & AUPRC & SD & Macro-F1 & AUROC & Status note \\
\midrule
\endhead
Amazon & Feature MLP & 5 & 0.8845 & 0.0087 & 0.9197 & 0.9780 & complete \\
Amazon & One-hop mean-aggregation MLP & 5 & 0.8761 & 0.0141 & 0.9065 & 0.9763 & complete \\
Amazon & BPM structural & 5 & 0.8765 & 0.0127 & 0.9057 & 0.9779 & complete \\
Amazon & Neighbor risk & 5 & 0.3214 & 0.0326 & 0.6540 & 0.7848 & complete \\
Amazon & EB neighbor evidence & 5 & 0.2995 & 0.0241 & 0.6538 & 0.8323 & complete \\
Amazon & Label-free context & 5 & 0.9085 & 0.0080 & 0.9227 & 0.9843 & complete \\
Amazon & Evidence context & 5 & 0.7715 & 0.0157 & 0.8458 & 0.9564 & complete \\
Amazon & Exploratory full gate & 5 & 0.9050 & 0.0088 & 0.9214 & 0.9839 & complete \\
\rowcolor{prefshade} Amazon & PREF-Gate (adaptive gate) & 5 & 0.9085 & 0.0080 & 0.9227 & 0.9843 & complete \\
Amazon & CARE-GNN & 5 & 0.8422 & 0.0130 & 0.9099 & 0.9570 & complete \\
Amazon & PC-GNN & 5 & 0.8097 & 0.0303 & 0.8762 & 0.9509 & complete \\
Amazon & BWGNN & 5 & 0.8601 & 0.0084 & 0.9142 & 0.9754 & complete \\
Amazon & PMP & 5 & 0.8690 & 0.0124 & 0.9129 & 0.9755 & complete \\
Amazon & ConsisGAD & 5 & 0.8696 & 0.0186 & 0.9180 & 0.9753 & complete \\
\addlinespace[2pt]
YelpChi & Feature MLP & 5 & 0.5245 & 0.0091 & 0.7105 & 0.8350 & complete \\
YelpChi & One-hop mean-aggregation MLP & 5 & 0.6136 & 0.0072 & 0.7501 & 0.8642 & complete \\
YelpChi & BPM structural & 5 & 0.6131 & 0.0072 & 0.7487 & 0.8635 & complete \\
YelpChi & Neighbor risk & 5 & 0.2757 & 0.0073 & 0.5765 & 0.6356 & complete \\
YelpChi & EB neighbor evidence & 5 & 0.2670 & 0.0067 & 0.5748 & 0.6391 & complete \\
YelpChi & Label-free context & 5 & 0.8104 & 0.0043 & 0.8405 & 0.9438 & complete \\
YelpChi & Evidence context & 5 & 0.5932 & 0.0071 & 0.7323 & 0.8654 & complete \\
YelpChi & Exploratory full gate & 5 & 0.8090 & 0.0062 & 0.8396 & 0.9436 & complete \\
\rowcolor{prefshade} YelpChi & PREF-Gate (adaptive gate) & 5 & 0.8104 & 0.0043 & 0.8405 & 0.9438 & complete \\
YelpChi & CARE-GNN & 5 & 0.3813 & 0.0022 & 0.6447 & 0.7653 & complete \\
YelpChi & PC-GNN & 5 & 0.4925 & 0.0048 & 0.6998 & 0.8184 & complete \\
YelpChi & BWGNN & 5 & 0.5105 & 0.0095 & 0.7017 & 0.8275 & complete \\
YelpChi & PMP & 5 & 0.7340 & 0.0179 & 0.8012 & 0.9252 & complete \\
YelpChi & ConsisGAD & 5 & 0.5397 & 0.0140 & 0.7198 & 0.8605 & complete \\
\addlinespace[2pt]
TFinance & Feature MLP & 5 & 0.7211 & 0.0158 & 0.8352 & 0.9386 & complete \\
TFinance & One-hop mean-aggregation MLP & 5 & 0.8101 & 0.0113 & 0.8842 & 0.9496 & complete \\
TFinance & BPM structural & 5 & 0.8075 & 0.0137 & 0.8839 & 0.9488 & complete \\
TFinance & Neighbor risk & 5 & 0.8183 & 0.0108 & 0.8730 & 0.9470 & complete \\
TFinance & EB neighbor evidence & 5 & 0.8193 & 0.0109 & 0.8741 & 0.9508 & complete \\
TFinance & Label-free context & 5 & 0.8860 & 0.0090 & 0.9204 & 0.9647 & complete \\
TFinance & Evidence context & 5 & 0.8350 & 0.0071 & 0.8830 & 0.9631 & complete \\
TFinance & Exploratory full gate & 5 & 0.8913 & 0.0106 & 0.9212 & 0.9724 & complete \\
\rowcolor{prefshade} TFinance & PREF-Gate (adaptive gate) & 5 & 0.8913 & 0.0106 & 0.9212 & 0.9724 & complete \\
TFinance & CARE-GNN & 5 & 0.7259 & 0.0110 & 0.8598 & 0.9146 & complete \\
TFinance & BWGNN & 5 & 0.7648 & 0.1922 & 0.8538 & 0.9440 & complete \\
TFinance & ConsisGAD & 5 & 0.8668 & 0.0090 & 0.9057 & 0.9631 & complete \\
\bottomrule
\end{longtable}
\normalsize

\end{landscape}

Three points are visible in the complete table. First, neither raw
neighbor-risk probability nor the empirical-Bayes posterior mean is competitive
as a stand-alone detector; evidence becomes useful only when combined with
attribute and graph context. Second, the label-free context expert is the
dominant internal component on Amazon and YelpChi. Third, the adaptive gate
matches the highest-scoring internal member within the present evaluation on
all datasets while using an evidence mixture only on TFinance.

\clearpage
\section{Split-Level Adaptive Results}
\label{app:splits}

Table~\ref{tab:per-split} exposes every adaptive-gate result used to compute the
means and standard deviations. Reporting the rows avoids a common ambiguity in
which a paper provides only aggregate values but does not permit paired
reconstruction. The five rows per dataset align by outer seed with all baseline
rows.

\small
\renewcommand{\arraystretch}{1.08}
\begin{longtable}{lrrcccc}
\caption{Split-level PREF-Gate results.}\label{tab:per-split}\\
\toprule
Dataset & Seed & Precision@5\% & Recall@5\% & AUPRC & Macro-F1 \\
\midrule
\endfirsthead
\toprule
Dataset & Seed & Precision@5\% & Recall@5\% & AUPRC & Macro-F1 \\
\midrule
\endhead
Amazon & 11 & 0.9707 & 0.7052 & 0.9164 & 0.9300 \\
Amazon & 23 & 0.9623 & 0.6991 & 0.9022 & 0.9167 \\
Amazon & 37 & 0.9665 & 0.7021 & 0.9080 & 0.9232 \\
Amazon & 53 & 0.9456 & 0.6869 & 0.8991 & 0.9169 \\
Amazon & 71 & 0.9498 & 0.6900 & 0.9167 & 0.9268 \\
\addlinespace[2pt]
TFinance & 11 & 0.7627 & 0.8336 & 0.8739 & 0.9109 \\
TFinance & 23 & 0.7754 & 0.8474 & 0.8914 & 0.9262 \\
TFinance & 37 & 0.7893 & 0.8627 & 0.8955 & 0.9248 \\
TFinance & 53 & 0.7944 & 0.8682 & 0.9026 & 0.9226 \\
TFinance & 71 & 0.7779 & 0.8502 & 0.8929 & 0.9214 \\
\addlinespace[2pt]
YelpChi & 11 & 0.9630 & 0.3317 & 0.8168 & 0.8425 \\
YelpChi & 23 & 0.9533 & 0.3283 & 0.8077 & 0.8440 \\
YelpChi & 37 & 0.9533 & 0.3283 & 0.8111 & 0.8417 \\
YelpChi & 53 & 0.9576 & 0.3298 & 0.8055 & 0.8367 \\
YelpChi & 71 & 0.9565 & 0.3295 & 0.8111 & 0.8376 \\
\bottomrule
\end{longtable}
\normalsize

The split-level AUPRC range is narrow on YelpChi (0.8055--0.8168) and Amazon
(0.8991--0.9167). TFinance varies from 0.8739 to 0.9026, but the adaptive gate
still exceeds ConsisGAD on all five paired splits. No seed is removed from an
aggregate.

\clearpage
\section{Frozen Hyperparameters and Selection Rules}
\label{app:hyperparameters}

\begin{table}[t]
\centering
\caption{Frozen PREF-Gate configuration.}
\label{tab:hyperparameters}
\resizebox{\textwidth}{!}{%
\begin{tabular}{lll}
\toprule
Component & Field & Value \\
\midrule
Label-free context & Input & $x_i$, neighbor mean, absolute residual, log degree, availability \\
 & Classifier & Histogram gradient boosting \\
 & Iterations / learning rate & 180 / 0.05 \\
 & Maximum leaf nodes / $L_2$ & 31 / 1.0 \\
 & Class weighting / seed & balanced / outer seed $+303$ \\
\midrule
Evidence context & Additional input & raw risk, posterior mean/variance, support, availability, displacement, prior \\
 & Classifier & Extra-Trees \\
 & Trees / minimum leaf & 200 / 10 \\
 & Class weighting / seed & balanced / outer seed $+404$ \\
\midrule
Empirical Bayes & Concentration bounds & $[2,100]$ \\
 & Fallback concentration & 12 \\
 & Label source & outer-training labels only, with target self-exclusion \\
\midrule
Adaptive gate & Context weights & $\{0,0.25,0.50,0.75,1\}$ across five candidates \\
 & Selection score & $0.5$ validation Macro-F1 $+0.5$ validation AUPRC \\
 & Threshold & validation Macro-F1 optimum \\
 & Tie breaking & AUPRC, Macro-F1, candidate name \\
\bottomrule
\end{tabular}}
\end{table}

The selected threshold is not constrained to 0.5 because class imbalance and
the different probability scales of the two experts make a universal threshold
inappropriate. Threshold selection is repeated independently within each
outer validation set, but the rule itself is fixed.

\clearpage
\section{Recent-Baseline Screening Boundary}
\label{app:screening}

\begin{table}[t]
\centering
\caption{Recent-method screening decisions. ``Quantitative'' means that a
same-split, label-safe, deferred-test adapter produced final metrics.}
\label{tab:screening}
\resizebox{\textwidth}{!}{%
\begin{tabular}{llll}
\toprule
Method & Venue/year & Decision & Principal reason \\
\midrule
PMP & ICLR 2024 & Quantitative & Passed adapter; TFinance host-memory failure disclosed \\
ConsisGAD & ICLR 2024 & Quantitative & Passed adapter on all three datasets \\
DGA-GNN & AAAI 2024 & Related work only & Pinned stack absent; stock test loop not admissible \\
DiG-In-GNN & AAAI 2024 & Related work only & Internal split and unconditional test path \\
GAAP~\cite{duan2025gaap} & AAAI 2025 & Related work only & Released local configurations do not complete the common matrix \\
CGNN~\cite{li2025cgnn} & AAAI 2025 & Related work only & Extremely limited-label protocol differs from 40/20/40 supervision \\
HUGE~\cite{pan2025huge} & AAAI 2025 & Related work only & Unsupervised task differs from supervised protocol \\
MimbFD~\cite{song2025mimbfd} & IJCAI 2025 & Related work only & No passing frozen-split deferred-test adapter in this workspace \\
ARC & NeurIPS 2024 & Related work only & In-context/few-shot task is not interchangeable \\
\bottomrule
\end{tabular}}
\end{table}

The screening rule was methodological, not outcome based. A recent method was
not excluded because its reported result was high; it was excluded when the
current workspace could not produce a traceable result under the same
information boundary. Conversely, established older methods remain because
their model cores could be evaluated fairly.

\clearpage
\section{Reproducibility Checklist}
\label{app:checklist}

\begin{itemize}
  \item \textbf{Data identity:} dataset artifacts and normalized graph policies
  are recorded.
  \item \textbf{Split identity:} every row contains train, validation, test, and
  combined index hashes.
  \item \textbf{Method identity:} source-bundle and method-configuration hashes
  are stored per row.
  \item \textbf{Selection identity:} all five validation candidates, their
  thresholds, and their ranking are preserved.
  \item \textbf{Prediction identity:} compressed prediction artifacts have
  SHA256 hashes and include test-node order.
  \item \textbf{Failure identity:} method and infrastructure failures remain
  explicit in the status matrix.
  \item \textbf{Statistical identity:} paired tests are reconstructed from
  outer-split rows, not from pooled node predictions.
  \item \textbf{Resume behavior:} completed rows are not executed again.
\end{itemize}

The checklist does not eliminate all reproducibility risks. External source
repositories, package versions, and public dataset hosting can change. It does,
however, make the final evidence internally auditable and prevents accidental
mixing with earlier development or unrelated project result tables.

\section*{Data Availability}

Amazon and YelpChi are available through the public DGL fraud dataset
interface, and TFinance is used under its released access terms. Split hashes,
aggregate and split-level metrics, statistical tests, source hashes,
prediction artifacts, and reproducibility manifests are retained in the
author-verified research record. Code, split manifests, and processed
evaluation artifacts will be released through an author-verified repository
upon acceptance. Raw datasets will not be redistributed.

\section*{Declaration of Competing Interest}

The authors declare that they have no known competing financial interests or
personal relationships that could have appeared to influence the work
reported in this paper.

\bibliographystyle{elsarticle-num}
\bibliography{references}
\end{document}